\documentclass[runningheads]{llncs}

\usepackage{eccv}

\usepackage{eccvabbrv}

\usepackage{graphicx}
\usepackage{booktabs}

\usepackage{subcaption}

\usepackage{multirow}

\usepackage[accsupp]{axessibility}  

\usepackage[pagebackref,breaklinks,colorlinks,citecolor=eccvblue]{hyperref}

\usepackage{orcidlink}

\begin{document}

\title{Few-shot Novel View Synthesis using Depth Aware 3D Gaussian Splatting} 

\titlerunning{Few-shot NVS using Depth Aware 3D Gaussian Splatting}


\author{
  Raja Kumar\thanks{Authors contributed equally} \and
  Vanshika Vats$^\star$\orcidlink{0000-0002-0967-6838}
}

\authorrunning{Kumar and Vats}

\institute{University of California Santa Cruz, CA 95064, USA \newline
\email{\{rkumar38,vvats\}@ucsc.edu}}

\maketitle

\begin{abstract}
  3D Gaussian splatting has surpassed neural radiance field methods in novel view synthesis by achieving lower computational costs and real-time high-quality rendering. Although it produces a high-quality rendering with a lot of input views, its performance drops significantly when only a few views are available. In this work, we address this by proposing a depth-aware Gaussian splatting method for few-shot novel view synthesis. We use monocular depth prediction as a prior, along with a scale-invariant depth loss, to constrain the 3D shape under just a few input views. We also model color using lower-order spherical harmonics to avoid overfitting. Further, we observe that removing splats with lower opacity periodically, as performed in the original work, leads to a very sparse point cloud and, hence, a lower-quality rendering. To mitigate this, we retain all the splats, leading to a better reconstruction in a few view settings. Experimental results show that our method outperforms the traditional 3D Gaussian splatting methods by achieving improvements of 10.5\% in peak signal-to-noise ratio, 6\% in structural similarity index, and 14.1\% in perceptual similarity, thereby validating the effectiveness of our approach. The code will be made available at \href{https://github.com/raja-kumar/depth-aware-3DGS}{https://github.com/raja-kumar/depth-aware-3DGS}.
  
\end{abstract}

\section{Introduction}
\label{sec:intro}

Novel View Synthesis (NVS), where the goal is to render unseen viewpoints given a set of input images, is a long-standing problem in the field of 3D vision. Neural Radiance Field (NeRF) \cite{mildenhall2021nerf} emerged as a state-of-the-art representation to achieve the highest quality renderings. Although NeRF achieves high-quality rendering, it suffers from high computational overhead due to the use of a Multi-Layer Perceptron (MLP) and differential volumetric rendering \cite{ niemeyer2022regnerf, mildenhall2021nerf, yang2023freenerf}. Recently, Kerbl et al. \cite{kerbl20233d} introduced a 3D Gaussian based representation with point-based differentiable rendering, achieving a very high-quality and real-time rendering. 


Although this method achieves great results, it relies on many input views and dense coverage of the scene for a faithful reconstruction. However, in real-world applications such as AR/VR, autonomous driving, and robotics, the number of available views for any object is few and sparse. In such scenarios, the rendering quality of the existing methods drops significantly. Existing works try to address this challenge by using various regularizations on geometry and appearances \cite{niemeyer2022regnerf, Jain_2021_ICCV, yang2023freenerf}. In our study, we propose a few-shot novel view synthesis using depth-aware 3D Gaussian splatting. We achieve this by using off-the-shelf monocular depth prediction as a prior to constrain the 3D shape of a scene. The original method uses high-order spherical harmonics (SH) to represent the color of each point which helps them recover high-frequency details. However, it is difficult to recover high-frequency intricacies given only a few input views since it leads to over-fitting the training views. Therefore, we model the color using lower-order spherical harmonics. Our experimental results show that the proposed depth-aware approach with just 5 views outperforms the existing state-of-the-art 3DGS method under few-shot conditions.

Thus, the main contribution of our study is synthesizing good quality novel views from a very limited data to begin with, using depth-aware 3D Gaussian splatting and achieving state-of-the-art performance on PSNR, SSIM and LPIPS metrics.



\section{Related Works}
\label{sec:relatedworks}

\subsection{Neural Representations of Scenes}

Coordinate-based Neural Representation \cite{chen2019learning, mescheder2019occupancy, michalkiewicz2019implicit} has gained popularity in recent years. Mildenhall et al. \cite{mildenhall2021nerf} introduced NeRF, an MLP-based scene representation that predicts the density and color of a 3D point given its coordinate and view direction. It achieved state-of-the-art results in synthesizing novel views of complex scenes. This work laid the foundation of several studies applying neural radiance fields in various domains including model reconstruction \cite{oechsle2021unisurf, yariv2021volume}, video representations \cite{nirvana, li2022neural, chen2023hnerv}, pose estimation \cite{Su2021ANeRFSH, chen2022dfnet} and editing \cite{kobayashi2022distilledfeaturefields, hyung2023local}. The main challenge of the NeRF based method lies in the heavy computation cost requiring hours of training even for a single scene. The challenge arises from the large parameter MLP and a slow volumetric rendering process. To improve the training time, M\"uller et al. \cite{mueller2022instant} proposes a Multiresolution Hash Encoding-based representation which avoids the MLP and also provides a faster rendering leading to drastic improvement in the training time. FastNeRF \cite{garbin2021fastnerf} uses graphics-inspired factorization to compactly cache a deep radiance map at each position in space and efficiently query that map using ray directions to improve the rendering speed of NeRF. To further improve the rendering quality and avoid the aliasing issue of NeRF, Mip-NeRF \cite{barron2021mip} replaces the point-based ray tracing using anti-aliased conical frustums. 

A recent breakthrough in NVS was achieved when Kerbl et al. \cite{kerbl20233d} proposed a 3D Gaussian representation with a point-based rendering that trains a scene within a few minutes and a rendering speed of more than 100FPS at 1080p resolution. In this method, each scene is represented as a set of 3D Gaussians with a size and orientation. A point-based differentiable rendering method inspired by Zwicker et al. \cite{zwicker2001surface} helps them achieve a real-time rendering speed. Expanding the applicability of 3DGS to dynamic scenes, Liuten et al. \cite{luiten2023dynamic} presented an innovative approach that allows 3D Gaussians to move and rotate over time, facilitating dynamic scene novel-view synthesis alongside six-degrees-of-freedom (6-DOF) tracking. Furthering the research, Yu et al.\cite{yu2023cogs} proposed CoGS, a controllable Gaussian Splatting method that enables real-time manipulation of dynamic scenes. Yang et al.\cite{yang2023deformable3dgs} introduced deformable 3D Gaussian Splatting, which reconstructs scenes with explicit 3D Gaussians and a deformation field. Notable other works have also made 3D Gaussian Splatting their primary choice for novel view synthesis given their faster implementation times and good quality rendering \cite{zheng2023gpsgaussian, liang2023gaufre, zhou2024headstudio}.

Although these methods improve both the rendering quality and computational cost, they still require many views to reconstruct any scene. In our work, we use 3D Gaussian based representation to represent the scene using a few input views.

\begin{figure*}[t]
  \centering
  \includegraphics[clip, trim=1cm 6cm 0.5cm 4.7cm, width=1\linewidth]{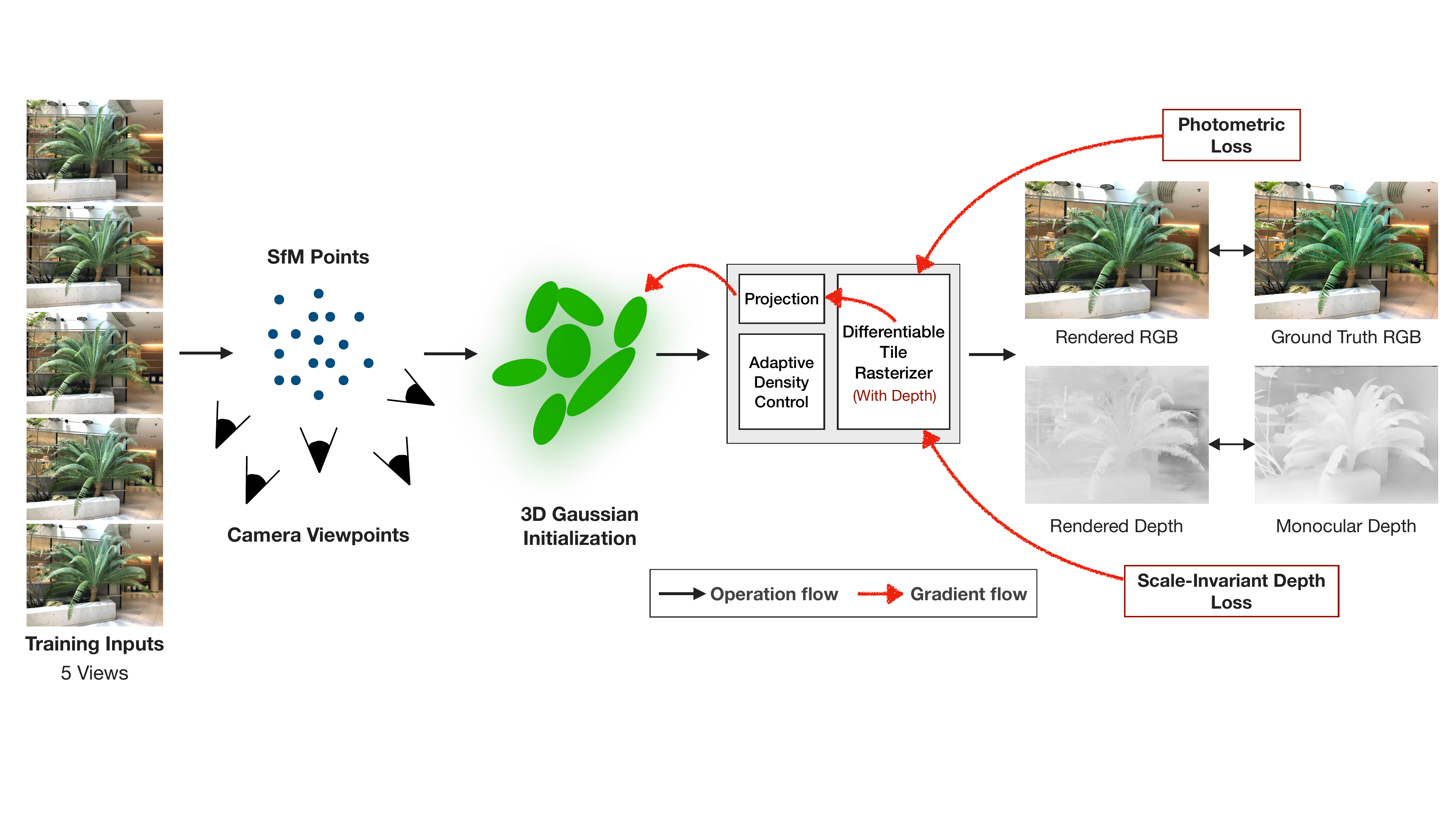}
  \caption{An overview of our proposed method. We start with just 5 training inputs from different viewpoints, collect Structure-from-Motion (SfM) points, and initialize the 3D Gaussians from them. We then use our modified differentiable tile rasterizer to render depth. In addition to photometric loss, we also compute scale-invariant depth loss and use it as supervision.}
  \label{figure:method}
\end{figure*}

\subsection{Few-Shot Novel View Synthesis}

One of the common approaches to solving the problem of few-views reconstruction is to use a prior to constraint the optimization \cite{kumar2023disjoint}. Several works \cite{niemeyer2022regnerf, Jain_2021_ICCV, yang2023freenerf, prinzler2023diner} have tried to achieve few-shot reconstruction using NeRF representation. Niemeyer et al. \cite{niemeyer2022regnerf} overcome this issue of NeRF's performance degradation with sparse input views by regularizing the geometry and appearance of rendered patches and using a normalizing flow model to regularize color. Jain et al. \cite{Jain_2021_ICCV} use semantic consistency loss to generate realistic novel renderings from only a few views. Yang et al. \cite{yang2023freenerf} propose a simple frequency range and near-camera density field regularization to improve the performance of NeRF under few views. Prinzler et al. \cite{prinzler2023diner} proposed to use a multi-view depth prediction as supervision during NeRF training. Taking inspiration from this work, we try to apply a similar approach to 3D Gaussian Splatting (3DGS). Although these methods achieves better rendering quality using few views input images, it suffers from the inherent limitation of NeRF i.e high computational cost due to volumetric rendering.


To reduce the training time even further, the next step in this direction would be to implement Gaussian splatting with limited views. Since 3DGS is a relatively newer concept in NVS, there is limited related work in few-shot 3DGS. We only found a few concurrent unpublished works \cite{xiong2023sparsegs, zhu2023FSGS} in a similar line as ours. Our study, however, is developed independently from the other works and in a similar timeframe. Therefore, to the best of our knowledge, our study excels over the other published works in the few-shot novel view synthesis by making 3DGS depth-aware on just 5 training views and gaining a good performance increase over the original methods.



\section{Methodology}
\label{sec:methods}

\begin{figure}[ht!]
  \centering
  \includegraphics[clip, trim=0cm 0cm 2cm 0cm, width=0.9\linewidth]{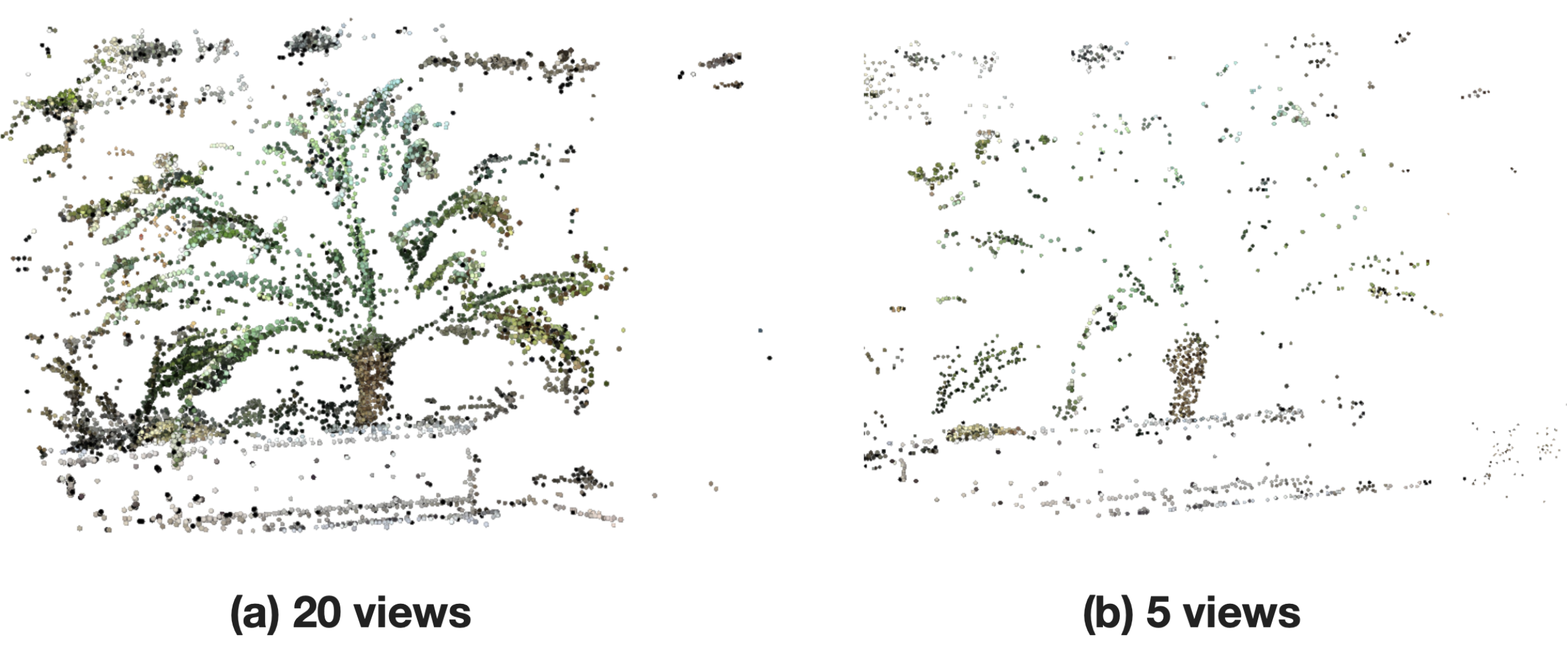}
  \caption{Structure from Motion (SfM) point clouds extracted using COLMAP \cite{schoenberger2016sfm} from (a) original 20 views, and (b) 5 views. Notice how sparse the SfM points are for 5 views with which we initiate our training.}
  \label{fig:sfm}
\end{figure}

In this section, we discuss our proposed method in detail. Fig.  \ref{figure:method} shows the overall pipeline of our method. Given 5 input views, we use COLMAP \cite{schoenberger2016sfm, schoenberger2016mvs} to get the initial point cloud and camera parameters. Using this sparse point cloud, we initialize our 3D Gaussians for training. As shown in Fig. \ref{fig:sfm}, the point cloud generated using Structure From Motion (SFM) from 5 views is very sparse compared to the one generated using many views. After 3D Gaussian initialization, we perform the projection and adaptive density control followed by rendering to get the rendered image. We modify the original renderer to render depth in addition to the RGB image. We use the implementation provided in \cite{kiuiDepth} for rendering depth. Once we have the rendered image and depth, we compute the loss. In the original 3DGS method, pixel-wise photometric loss, a combination of L1 and SSIM loss, is used to train the model. However, in the case of few views, this photometric loss alone is not enough as it tends to overfit the training views while the 3D shape is still inaccurate. To this end, we propose to use a depth prior to constrain the 3D shape. In the following sections, we first discuss the details of the original 3D Gaussian splatting method, followed by our depth prior and scale-invariant depth loss regularization term.


\subsection{3D Gaussian Splatting}
3DGS, originally proposed by Kerbl et al.\cite{kerbl20233d}, represents a scene using 3D Gaussians characterized by position, covariance, opacity, and spherical harmonics for color. The optimization of this compact scene representation involves adjusting these parameters and controlling Gaussian density. Efficiency is achieved through a tile-based rasterizer that blends anisotropic splats in visibility order, supporting a vast number of Gaussians for gradient reception without limits. 

An obvious approach would be to optimize the covariance matrix $\Sigma$. However, covariance matrices have physical meaning only when they are positive semi-definite. However, it is challenging to constrain the optimization to generate such matrices. Therefore, it is represented using a more intuitive representation of orientation. Given a scaling matrix $S$ and rotation matrix R, the corresponding $\Sigma$ can be computed as 

\begin{equation}
    \Sigma = RSS^TR^T
    \label{equation1}
\end{equation}

In addition, Kerbl et al. \cite{kerbl20233d} also use an optimization with adaptive density control to adaptivly control the number of Gaussians and their density over unit volume, allowing the Gaussian splats to go from an initial sparse set of Gaussians to a denser set that better represents the scene, and with correct parameters. This optimization basically adjusts the Gaussian distribution to fill voids in scenes, targeting areas lacking geometric detail or overly broad Gaussian coverage, indicative of under- or over-reconstruction. Such regions are identified by significant view-space positional gradients, suggesting they're not fully optimized, prompting the algorithm to reposition Gaussians for better accuracy.


\subsection{Depth Regularization}
The original 3DGS \cite{kerbl20233d} algorithm produces its best results using many viewpoints. However, the rendering quality decreases significantly when only a few views are available. This loss in quality is mainly due to the model fitting the training views without learning the intrinsic 3D shape correctly. With the advent of deep learning, existing works \cite{ranftl2021vision} can predict the depth of a scene from a single view accurately. In this work, we propose to use this depth prior to the few-shot 3DGS learning to constrain the shape.

\subsubsection{Gaussian Splatting Depth Rendering}
3DGS uses a neural point-based approach to render the color of pixels. The color $C$ of a pixel is computed by blending $N$ ordered points overlapping the pixel as:

\begin{equation}
    C = \sum_{i \in N} {c_i \alpha_i}{{\prod_{j=1}^{i-1} (1 - \alpha_j)}}
    \label{equation2}
\end{equation}

 where $c_i$ is the color of each point and $\alpha_i$ is the opacity of the projected 2D splats computed by multiplying the covariance $\sum$ with opacity. In order to use depth prior as supervision, we need to modify the renderer to render the depth as well. To this end, we use differential Gaussian rasterization \cite{kiuiDepth}, which implements depth and alpha rendering in addition to color. Similar to pixel rendering, it uses alpha blending of z-buffer from the ordered Gaussians. The depth $D$ of a point is computed as:

 \begin{equation}
     D = \sum_{i \in N} {d_i \alpha_i}{{\prod_{j=1}^{i-1} (1 - \alpha_j)}}
     \label{equation3}
 \end{equation}
 where $d_i$ is the z-buffer value for $i^{th}$ gaussian and $\alpha_i$ is same as equation \ref{equation2}.

\subsubsection{Scale Invariant Depth Loss}
Given the depth estimated from the monocular depth estimation model and the rendered depth, we can calculate the loss between these two for supervision. One of the challenges of calculating the depth loss is the scale of depth values. The scale of the monocular model and the 3DGS depth renderer can be different. Therefore, based on Eigen et al.'s \cite{eigen2014depth} work, we use a scale-invariant loss function that takes into account the scale of both monocular and 3DGS depth renderer. From a rendered depth map $y$ and predicted depth (using monocular model) $y^*$
, each with n pixels index by i, we compute scale invariant mean squared error (in log space) as 

\begin{equation}
    L_{depth}(y, y^*) = \frac{1}{2n} \sum_{i=1}^{n} \left( \log y_i - \log y^*_i + \alpha(y, y^*) \right)^2
\end{equation}
where, $\alpha(y, y^*) = \frac{1}{n} \sum_{i} \left( \log y^*_i - \log y_i \right)$ is the value of $\alpha$ that minimizes the error for a give $(y, y^*)$.

In addition to the photometric pixel-wise loss used in the original 3DGS paper, we use this scale invariant depth loss. Therefore our final loss function would become 

\begin{equation}
    L = (1-\lambda)L_1 + \lambda L_{D-SSIM} + \lambda_{depth}L_{depth}
\end{equation}

We set the value of $\lambda = 0.2$ and $\lambda_{depth} = 0.005$ for our tests.

\subsection{Other Optimizations}

In addition to using monocular depth prior, we also make modifications to the Spherical Harmonics (SH) model and density pruning step to make it fit for few view settings. In the original 3DGS, a SH coefficient of order 3 is used to estimate the color of each point. A higher order harmonics is basically fitting a higher degree polynomial \cite{ramamoorthi2006modeling}, and hence a more detailed reconstruction can be achieved. However, when we only have a few views available for training, it is challenging to reconstruct high-frequency details. Based on this observation, we decrease our SH coefficient order to 1, making our scene less prone to overfitting. In the original 3DGS, the Gaussians with low opacity are removed periodically. However, we observe that removing splats with lower opacity periodically leads to removing a large amount of the splats since the initial point cloud is very sparse in a few view inputs. This leads to an unstable and inaccurate optimization. We address this by retaining all the splats, which leads to a better reconstruction in a few view settings.


\section{Experiments}
\label{sec:experiments}


Here, we first discuss the datasets, evaluation
metrics, and implementation details of the experiments. We then perform a qualitative and quantitative
comparison to the original 3DGS method to show the effectiveness of our method. 


\begin{figure*}[t]
  \centering
  \includegraphics[clip, trim=0cm 3cm 0cm 0cm, width=0.95\linewidth]{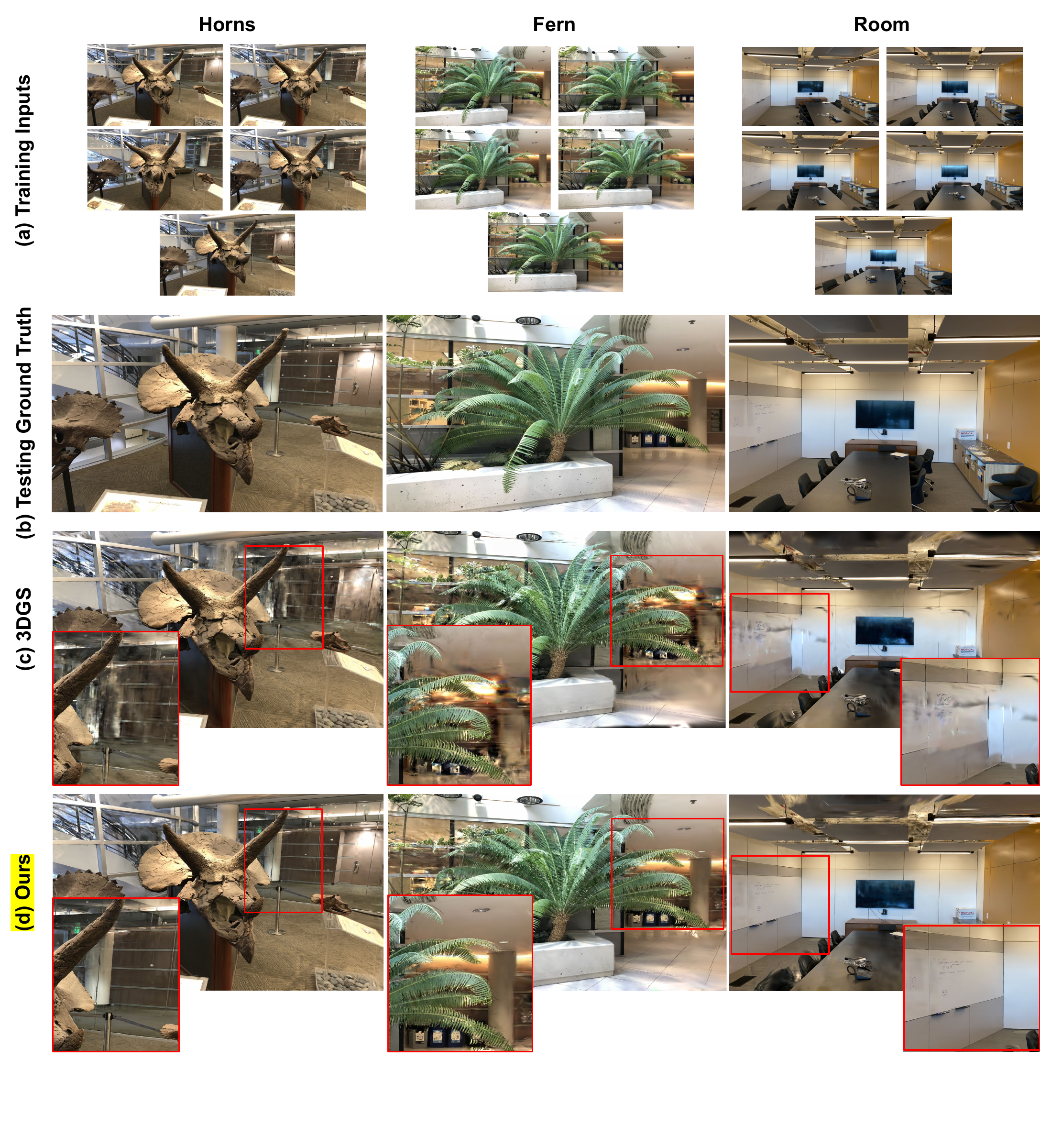}
  \caption{We compare our results with the original 3DGS and observe better rendering quality using our method. The red box shows zoomed-in output for better reference.}
  \label{fig:qualitative}
\end{figure*}

\subsection{Implementation Details}
We perform our experiments on the NeRF LLFF dataset \cite{LLFF}, which is widely acclaimed for NVS tasks. The original dataset contains 15-20 viewpoints for each scene. For our experiments, we sample 5 train and 3 test views uniformly, containing both interpolated and extrapolated viewpoints. By interpolated, we mean a view from a point in space that lies between the training camera positions, and by extrapolated, we assess the viewpoints extending beyond the range of the original camera position where no direct visual information is available from the original images. This is done to assess the robustness of our method under the challenging viewpoints. We set the order of spherical harmonics (SH) to 1 instead of the original 3, owing to the fact that it is challenging to model high-frequency details using only a few views. Hence a lower order SH can better represent the color. Also, as explained in Sec. \ref{sec:methods}, we do not reset the opacity of the splats periodically. We train each scene for 10000 iterations and report the widely used SSIM, PSNR, and LPIPS as evaluation metrics.

\begin{table*}[t]
\centering
\small
\resizebox{0.7\columnwidth}{!}{%
\begin{tabular}{@{}l|cc|cc|cc@{}}
\toprule
\multicolumn{1}{c|}{\multirow{2}{*}{\textbf{Scene}}} & \multicolumn{2}{c|}{\textbf{PSNR} $\uparrow$} & \multicolumn{2}{c|}{\textbf{SSIM} $\uparrow$} & \multicolumn{2}{c}{\textbf{LPIPS} $\downarrow$} \\ \cmidrule(l){2-7} 
\multicolumn{1}{c|}{} & 3DGS  & Ours           & 3DGS  & Ours           & 3DGS  & Ours           \\ \midrule
Room                   & 19.05 & \textbf{22.43} & 0.801 & \textbf{0.861} & 0.313 & \textbf{0.224} \\
Fern                   & 17.08 & \textbf{19.62} & 0.597 & \textbf{0.648} & 0.381 & \textbf{0.321} \\
Flower                 & 21.75 & \textbf{22.86} & 0.708 & \textbf{0.737} & 0.290 & \textbf{0.264} \\
Fortress               & 19.83 & \textbf{21.27} & 0.590 & \textbf{0.607} & 0.332 & \textbf{0.326} \\
Horns                  & 19.09 & \textbf{21.60} & 0.685 & \textbf{0.756} & 0.323 & \textbf{0.267} \\
Trex                   & 20.67 & \textbf{22.09} & 0.786 & \textbf{0.807} & 0.277 & \textbf{0.244} \\ \midrule
Average                & 19.57 & \textbf{21.64} & 0.694 & \textbf{0.736} & 0.319 & \textbf{0.274} \\ \bottomrule
\end{tabular}%
}
\vspace{0.3cm}
\caption{Quantitative comparison of the proposed method with the original 3DGS. Our depth-aware method consistently outperforms the original method on all three evaluation metrics across the scenes tested.}
\label{table:mainresults}
\end{table*}

\subsection{Experimental Results}

\label{results}
To evaluate the efficacy of our proposed method, we compare it against the original 3DGS method. We use the same few view settings to train the original 3DGS method and perform both quantitative and qualitative comparisons. 

Fig. \ref{fig:qualitative} demonstrates a qualitative comparison on three different scenes of the LLFF dataset. We show the five input views and rendered novel view using the original 3DGS method and our proposed method. We clearly see an improvement in the rendering quality. We highlight the zoomed-in region in the red box for better visualization and comparison. 

We also show the quantitative comparison on different scenes of the LLFF dataset in Table \ref{table:mainresults}. We see a clear improvement across all the metrics for each scene, with an average improvement of 10.5\%, 6.0\%, and 14.1\% in PSNR, SSIM, and LPIPS metrics, respectively. These improvements indicate that our method is able to learn the 3D representation better, and hence we also achieve a better rendering quality. 

\begin{figure*}[ht!]
  \centering
  \includegraphics[clip, trim=2.5cm 10cm 3cm 0cm, width=0.94\linewidth]{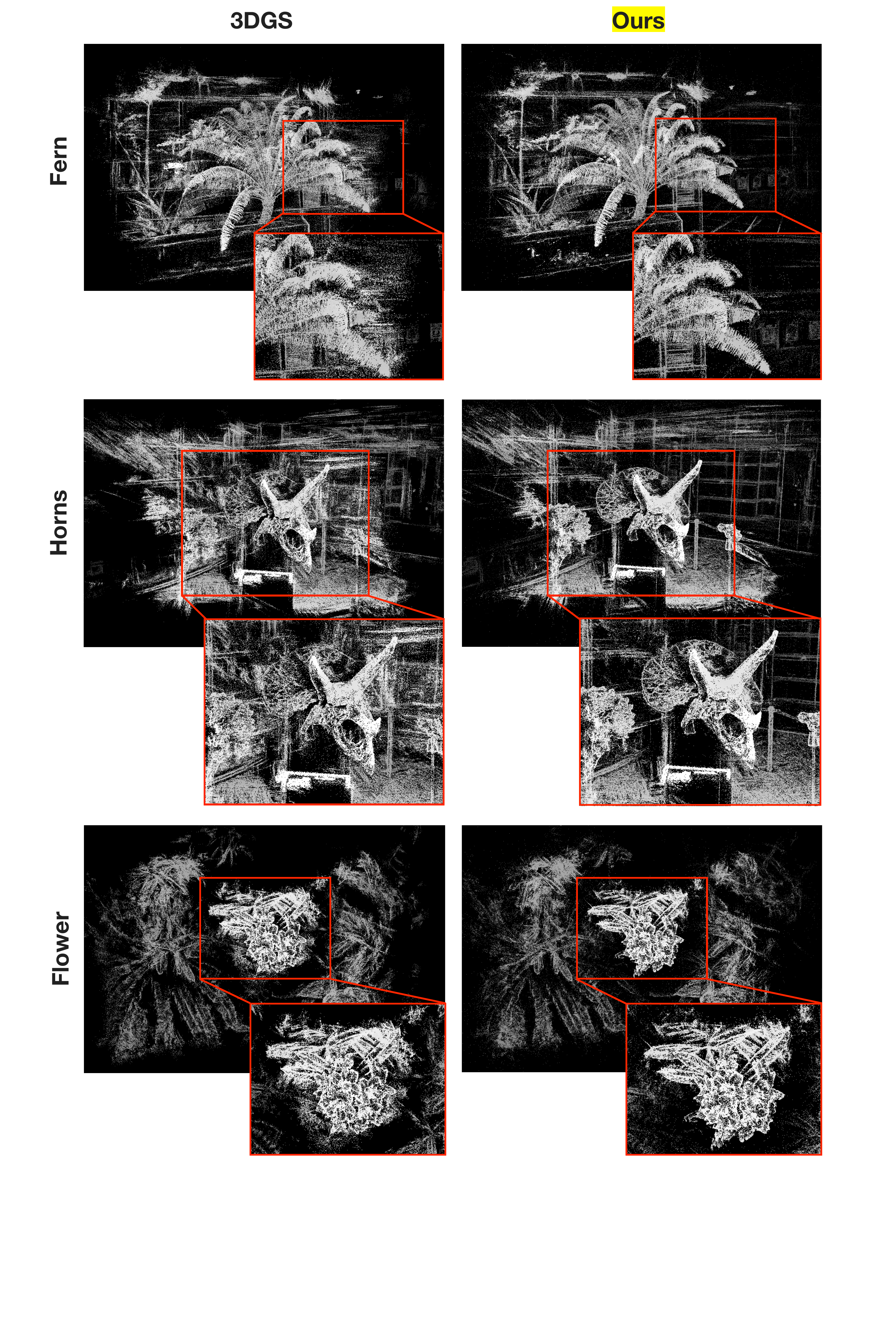}
  \caption{Visualization of point clouds generated from our method (right column) as compared to the original 3DGS (left column). Notice that our method forms cleaner and sharper point clouds than the original 3DGS. The red boxes show zoomed-in outputs for better reference and clarity.}
  \label{fig:pointcloud}
\end{figure*}

Given the depth data as a prior during training, we expect that the 3D shape should be improved compared to the training without depth. To this end, we compare the point cloud generated using our method and 3DGS in Fig.  \ref{fig:pointcloud}. The figure shows pseudo-depth maps on grayscale for visualization of point clouds. We can clearly see that our proposed method is able to recover the 3D shape better compared to the other method. The 3DGS point cloud has more noise and hence results in fuzzy renderings.

In our test set, we keep extrapolated viewpoints as well, which are challenging to synthesize. We hypothesize that since we are able to recover the 3D shape better, we must achieve better rendering for these viewpoints. We, therefore, compare the PSNR values of the extrapolated viewpoints in Table \ref{table:extrapolated}. We observe a good improvement across all the scenes. Overall, an improvement of 13.5\% is observed for these viewpoints indicating the efficacy and robustness of our method. Therefore, we can conclude that the proposed method is able to recover 3D shapes faithfully.




\section{Limitations and Future Work}
\label{sec:LimFut}

We achieve SOTA results on NVS with few input views. However, our method has limitations. Since we use an off-the-shelf monocular depth estimation method, our accuracy is dependent on the accuracy of the depth estimation model. Therefore, there remains a scope for improvement by using a better depth estimation model. In future work, Our method can be further improved by using a depth loss that matches the depth distribution rather than the absolute value. Another challenge of few-shot reconstruction is generating the initial point cloud for initialization. If the input views are too sparse or do not have enough matching points, the COLMAP reconstruction fails and hence the initialization. In future work, we plan to work on solving this by using a 3D bounding box prediction and then using random initialization. 

\section{Conclusion}
\label{sec:conclusion}

In this work, we propose a depth-aware 3D Gaussian splatting method for novel view synthesis that achieves state-of-the-art performance under few shot conditions. We use monocular depth estimation as a prior to supervise the training. Such a constraint helped us recover a more faithful 3D representation and, hence, a better rendering even for challenging viewpoints. We also make optimizations to the SH order and Gaussian density pruning technique to improve the rendering quality. Finally, We evaluate our method on the LLFF dataset, and our method outperforms the existing methods in terms of both quantitative and qualitative evaluation. 


\begin{table}[t]
\centering
\small
\resizebox{0.3\columnwidth}{!}{%
\begin{tabular}{@{}l|cc@{}}
\toprule
\multicolumn{1}{c|}{\multirow{2}{*}{\textbf{Scene}}} & \multicolumn{2}{c}{\textbf{PSNR} $\uparrow$} \\ 
\cmidrule(l){2-3}
                                    & 3DGS  & Ours           \\ \midrule
Room                                & 17.13 & \textbf{20.78} \\
Fern                                & 16.68 & \textbf{18.80} \\
Flower                              & 20.87 & \textbf{22.16} \\
Fortress                            & 12.59 & \textbf{16.68} \\
Horns                               & 17.60 & \textbf{19.43} \\
Trex                                & 19.71 & \textbf{20.81} \\ \midrule
Average                             & 17.42 & \textbf{19.78} \\ \bottomrule
\end{tabular}%
}
\vspace{0.3cm}
\caption{Quantitative comparison of the proposed method with the original 3DGS on the \emph{extrapolated} views. Our depth-aware method outperforms the original method on this hard-to-synthesize test set, proving the effectiveness of our method.}
\label{table:extrapolated}
\end{table}

\section{Acknowledgements}
We sincerely thank Prof. Yuyin Zhou and Prof. James Davis, UCSC, for their valuable feedback and insightful guidance.

%
%
\bibliographystyle{splncs04}
\bibliography{main}
\end{document}